# Adaptive Hoeffding Tree with Transfer Learning for Streaming Synchrophasor Data Sets

Zakaria El Mrabet
*School of Electrical Engineering and Computer Science (SEECS)*
*University of North Dakota*
Grand Forks, ND, USA
zakaria.elmrabet@und.edu

Daisy Flora Selvaraj
*School of Electrical Engineering and Computer Science (SEECS)*
*University of North Dakota*
Grand Forks, ND, USA
daisy.selvaraj@und.edu

Prakash Ranganathan
*School of Electrical Engineering and Computer Science (SEECS)*
*University of North Dakota*
Grand Forks, ND, USA
prakash.ranganathan@und.edu

*Abstract*— Synchrophasor technology or phasor measurement units (PMUs) are known to detect multiple type of oscillations or faults better than Supervisory Control and Data Acquisition (SCADA) systems, but the volume of Bigdata (e.g., 30-120 samples per second on a single PMU) generated by these sensors at the aggregator level (e.g., several PMUs) requires special handling. Conventional machine learning or data mining methods are not suitable to handle such larger streaming real-time data. This is primarily due to latencies associated with cloud environments (e.g., at an aggregator or PDC level), and thus necessitates the need for local computing to move the data on the edge (or locally at the PMU level) for processing. This requires faster real-time streaming algorithms to be processed at the local level (e.g., typically by a Field Programmable Gate Array (FPGA) based controllers). This paper proposes a transfer learning-based hoeffding tree with ADWIN (THAT) method to detect anomalous synchrophasor signatures. The proposed algorithm is trained and tested with the OzaBag method. The preliminary results with transfer learning indicate that a computational time saving of 0.7ms is achieved with THAT algorithm (0.34ms) over Ozabag (1.04ms), while the accuracy of both methods in detecting fault events remains at 94% for four signatures.

*Keywords— Hoeffding tree, ADWIN, Transfer learning, OzaBag, PMU, Oscillations.*

## I. Introduction

Recently, the U.S. power grid reported a cyber-attack incident that disrupted grid operations, specifically transmission operators in the western power region are hit by a Denial of Service (DoS) attack, causing a temporary loss of visibility in certain sections of the SCADA system. According to the report, "the hack itself happened on March 5th, 2019, when a denial of service attack disabled Cisco's adaptive security appliance devices, ringing power grid control systems in Utah, Wyoming, and California". this is a major and the first cyber-attack on the U.S power grid [1]. Other grid attacks include the Ukrainian power grid, which has experienced an outage for several hours impacting nearly 225,000 utility customers in 2015 due to cyber threats [2]. These two events illustrate the detrimental impacts of cyber-attacks and the economic burden that it can bring to any nation's critical infrastructure. To avoid such lack of visibility in the power grid, reliable sensor measurements are required such as Synchrophasor or PMU technologies [3].

The PMUs generally record the magnitude, phase angle, frequency, voltage and current phasors with a precise GPS based timestamp [4]. PMU data paired with real-time analytics can facilitate improved grid operations. In recent years, advanced machine learning (ML) methods are being deployed for mining multi-variate, large-scale PMU data to reveal new insights on anomalous events.

However, processing such large PMU streams requires expensive computational resources and faster algorithms. Further, the conventional ML algorithms [5]–[12], such as neural network, support vector machine, deep learning, and decision trees, cannot be used in this scenario, as they require loading and scanning the entire dataset before processing [13]. Thus, two important criteria have to be met by any streaming ML technique for handling PMU data: 1) training the model with recent history holding shorter signatures; and 2) adapting to concept drifts (e.g., fluctuations) in real-time [14]. It is important to note that the model trained on a historical record will no longer have relevance to the incoming data stream, and thus may fail to capture any critical or new events. Hence, meeting the above two criteria is key for successful and future real-time streaming algorithms [15], [16].

In [17], authors discuss a Hoeffding Tree (HT) combined with two concept drift detectors (e.g., drift detection method: (DDM) and Adaptive sliding windows: (ADWIN)) for building dynamic decision trees that are adaptable to quick changes. This method has been trained on a synthetic PMU dataset, where multiple attack scenarios were modeled. The dataset contains normal, anomalous, physical and cyber events. The physical event includes relay-based faults, and the cyber events include injections of various trip commands, single line to ground (SLG) fault replay commands and disable command attacks. The model (e.g., HT+ADWIN+DDM) trained and tested reported a classification accuracy greater than 98%.

Similarly, authors in [13] proposed a Hoeffding Adaptive Tree (HAT) based approach for detecting events on PMU data. The authors modeled two scenarios in their PMU data sets; 1) a three-phase fault has been generated with some load fluctuations altering true power (P) and reactive power (Q) at a regular interval; and 2) cataloged two classes such as fault and normal. Fault class includes a Single Line to Ground (SLG) fault, while the normal class includes normal power system variables such as voltage (v), current (i), frequency (f) and phase angle ($\phi$). Based on the reported results, Hoeffding tree showed a good ability in detecting the power system faults and adaptation to the concept drifts in comparison with traditional decision trees such as J48 and REPTree.

However, in the above two studies, the duration of the physical fault events was not considered in their data set. There are certain type of power system faults that are time-sensitive and thus require early detection. The cause of inter-area oscillations is primarily due to system events coupled with a poorly damped power system. Generally, these low-frequency oscillations (0.1-0.8 Hz) are noticed in a larger grid with multiple generators or renewable plants (with high wind or solar penetration) that are connected to weak tie line connections. This can lead to a high degree of uncertainties in the system and it is often difficult to detect in real-time.

Specifically, smaller frequency deviations that range from 0.15-1.0 Hz lasting 60 seconds or longer may cause inter-area oscillations and quickly destabilize the grid [18]. Similarly, momentary voltage or current instabilities (e.g., surges or spikes of 2-5 seconds) may lead to asset failures (e.g., transformer, relays, or circuit breakers). Thus, including signatures with various event duration is a key feature for training real-time machine learning algorithms.

In [19], the authors proposed a distributed Intrusion Detection System (IDS) using a multi-layer network architecture for Smart Grids. Here, IDSs are modeled at three levels of networks: 1) Home Area Network; 2) Neighborhood Area Network; and 3) Wide Area Network. These IDSs were based on the Support Vector Machine (SVM) and Artificial Immune system (AIS), where each classifier was trained and tested on the NSL_KDD dataset [20]. The results indicate that SVM outperformed AIS algorithms in terms of False Positive Rates (FPR) and False Negative Rates (FNR). However, the proposed classifiers are not suitable for streaming data, as they require loading the entire dataset for processing.

The authors of [21] have proposed an IDS for the Advanced Metering Infrastructure (AMI) data streams. The proposed IDS has been deployed at three levels of the AMI: Smart meter, AMI headend, and Data concentrator. They used seven data stream mining algorithms. They are as follows: 1) AccuracyUpdatedEnsemble; 2) ActiveClassifier; 3) LeveragingBag; 4) LimAttClassifier; 5) OzaBagAdwin; 6) OzaBagASHT; and 7) SingleClassifierDrift. These algorithms are trained and tested using the KDD Cup 1999 dataset [22]. Their findings indicate that both ActiveClassifier and SingleClassifierDrift algorithms reported satisfactory results in terms of saving memory and time. LeveraginBag provided higher accuracy and a lower false-positive rate, but it demands moderate memory. Thus, the method is an appropriate candidate for phasor data concentrators. For the data concentrator, which requires a response time of a fraction of second, ActiveClassifier can be used due to its faster processing time. However, one has to evaluate the number of available communication channels, and networks to validate the training scenarios and deployment of these algorithms.

The performance of the streaming data classifiers can further be improved by including pre-processing operations such as the removal of redundant data, and dimensionality reduction. A principal component analysis (PCA) was used for online dimensionality reduction for the PMU data stream [23]. This approach was tested on real PMU data generated by SEL412 and GE n60 PMUs, deployed on Real-Time Digital Power System Simulator (RTDS) in a 4-bus system. In [24], PCA was used for both detecting anomalies, and dimensionality reduction. Though the PCA methods appear to work well for dimensionality reduction, there is limited information on their ability to detect anomalous events. In addition, these works do not include concept drifts or signatures with varied duration and do not provide any comparison against existing techniques.

In this paper, a transfer learning-based Hoeffding Tree with ADWIN is proposed to detect anomalies in the streaming PMU data. Multiple HT classifier is trained on normal and anomalous signatures, while ADWIN is included at the leaf for adaptation to concept drift (e.g., fluctuations). Other concept drift detectors can be combined with the HT including drift detection method (DDM), Early Drift Detection Methods (EDDM), Linear Four Rate (LFR), Just in time (JIT), and ensemble methods [25], [26], [27]. DDM and EDDM don't require storing the data, but they are sensitive to false alarm rates. LFR [25] can deal with unbalanced classes, as the error types are handled differently. However, it suffers from the labeling cost. JIT [27] can detect the abrupt changes, and do not require labeled data, but cannot handle continuous and gradual drifts. Ensemble methods [28] are effective in detecting recurring concept drifts, however, it lacks in identifying the precise location of the drift, due to its larger batch size. In our work, we use ADWIN for identifying quick localized changes in the signatures. It also uses a dynamic window that adapts to varying size of the incoming streaming data. When data is stationary, the size of the window increases dynamically, and decreases, when a change is detected.

The proposed THAT model is trained on four event signatures with varying durations. As HT and ADWIN require fixed features during the training phase and they cannot be trained on events with different durations, these models will be improved by incorporating a transfer learning (TL) technique. Here, TL refers to the transfer (or retention) of knowledge that can be learned across multiple similar tasks, but not identical [29]. The process of TL is as follows: the model is trained on the first signature (e.g., first task) and any acquired knowledge will be transferred to perform the training on the next task, that focuses on the second signature. Thus, the process of transferring the learned training can be repeated to subsequent signatures. To the best of the author's knowledge, this is the first attempt to integrate the transfer learning with any streaming classifier for anomaly detection containing signatures with varying durations. In [29], the authors proposed a transfer learning in a decision tree, but it was not for streaming data applications. This paper is divided into the following sections: Section II covers an overview of THAT model, it details the training dataset, and the relevant features used for detecting the oscillation events; and section III evaluates the proposed model with some performance metrics. Section VI draws some conclusions.

## II. METHODOLOGY

### A. Transfer Adaptive Hoeffding tree (THAT)

The Hoeffding Tree (HT) is considered as a standard decision tree used for classification. It uses a Hoeffding bound that relies on the minimum number of arriving samples to build a certain confidence threshold to build trees. Therefore, avoids the entire data to be loaded into memory. The only information required is the algorithm itself, which stores enough statistics on incoming stream in its leaves, and thus enables the tree to grow, and classify its samples in real-time. Theoretically, if HT receives enough instances of data, then the method will have similar performance as conventional decision trees [14], [15]. Algorithm 1 illustrates the Transfer Hoeffding Adaptive tree (THAT) algorithm for the PMU/PDC stream. The algorithm has two stages, stage 1: create a new HT; and in stage 2: carry transfer learning operations between two HT models by training the target model on the subsequent signature in the queue using learned HT tree.

The details for these two stages are as follows: In stage 1, (lines 1-20), a new $HT_{target}$ tree is created if the $HT_{source}$ tree is not provided. During this early stage, transfer learning is not required. In line 2, a target tree is initialized by creating the first node (root). In lines 3-19, a for loop is performed for all the training instances, where each sample is filtered down the tree to the appropriate leaf $l$ based on the test sequences

**Algorithm 1**. Transfer learning Hoeffding Adaptive Tree (THAT)

Input: training set S, S', $HT_{source}$
Output: $HT_{target}$
1. If $HT_{source}$ is NULL
2.    Let $HT_{target}$ be a tree with a single leaf (the root)
3.    For all instances in $S$ do
4.      Sort instances into leaf $l$ using $HT_{target}$
5.      Update sufficient statistics in $l$
6.      Increment $n_l$, the number of instances seen at $l$
7.      If $n_l$ mod $n_{min} = 0$ and instances seen at $l$ not all of same class
8.        Compute $\overline{G_l}(A_i)$ for each attribute
9.        Let $A_a$ be attribute with highest $\overline{G_l}$
10.       Let $A_b$ be attribute with second highest $\overline{G_l}$
11.       Compute Hoeffding bound $\epsilon = \sqrt{\frac{R^2 \ln(\frac{1}{\delta})}{2n_l}}$
12.       If $A_a \neq A_0$ and $(\overline{G_l}(A_a) - \overline{G_l}(A_b)) > \epsilon$ or $\epsilon < \tau$)
13.         Replace $l$ with an internal node that splits on $A_a$
14.         For all branches of the split do
15.           Add a new leaf with sufficient statistics
16.         End for
17.       End if
18.      End if
19.    End for
20.    Return $HT_{target}$
21. Else
22.    $HT_{target} = HT_{source}$
23.    $Q \leftarrow$ all attributes of $S'$ not in $HT_{target}$
24.    For each attribute $A'$ dequeued from $Q$
25.      For each training instance $I$ in $S'$
26.        Classify $I'$ using the $HT_{target}$
27.        If $I'$ is predicted correctly then
28.         Do nothing
29.        Else
30.         Replace $HT_{target}$'s class node with a new node for attribute $A'$
31.         Add a new branch to node $A'$, labeled with $A$'s value in $I'$
32.         Add a new leaf node labeled with $I'$s target class label
33.        End if
34.      End for
35.    End for
36.    Return $HT_{target}$
37. End if

present in the $HT$ built to that point (line 4). During this process, sufficient statistics on the samples and Hoeffding bound is computed. Each leaf $l$ contains enough statistics in order to make decisions about the further growth of the tree. These statistics need to be sufficient to enable the calculation of the Information Gain afforded by each possible split. However, storing unnecessary information would increase the total memory requirement for the tree. In line 5, the statistics hold by $l$ are updated to estimate the Information Gain of splitting each attribute. In line 6, the $n_l$ number of samples seen at leaf $l$ (computed from the sufficient statistics), is updated. Lines 7-18 are executed only when a mix of different classes enables further splitting. In line 8, the splitting criterion $G$ is used to estimate the $\overline{G_l}$ value for each attribute. The function $G$ measures the average amount of purity that is gained in each subset of a split and indicates how well a given attribute separates the training examples according to their target classification [15]. In this paper, two different approaches will be investigated to compute the function $G$: Information Gain (entropy) and the Gini index. If the distribution of the two classes (e.g., oscillation event class, and normal event class) in the PMU stream contains the probabilities p1, and p2 of the classes, then the entropy of a given attribute $A$ in a training data set $S$ is calculated by:

$$Entropy\ (A) = \sum_{i=1}^{n} -p_i \log_2 p_i \quad (1)$$

Here $n$ is the number of classes and it is equal to 2. The attribute, $A$, is one of the selected features which could be voltage (V), frequency (f), current (I), or the phase angle (ϕ). Then, the Information Gain is computed by:

$$Information\ Gain\ (S, A) = Entropy(A) - \sum_{k \in (A)} \frac{|S_k|}{|S|} Entropy(S_k) \quad (2)$$

Where $k$ is the value of the attribute $A$, and $S_k$ is a subset of $S$ where $A = k$.

The other metric considered for evaluation is known as Gini Index, and this index is computed as:

$$Gini(A) = 1 - \sum_{i=1}^{n} p_j^2 \quad (3)$$

Here $n$ is the number of classes (e.g., n=2 in our case). In lines 9 and 10, the attributes with the largest gain information are used for the next steps. Line 11 computes the Hoeffding bound such that probability $1 - \delta$ corresponds to a confidence value of $\delta \in \{1,0\}$, where the true mean of a random variable of range $R$ will not differ from the estimated mean after $n$ independent observations by more than $\epsilon$ and it is given by:

$$\epsilon = \sqrt{\frac{R^2 \ln(\frac{1}{\delta})}{2n}} \quad (4)$$

In line 12, a split criterion test is performed between the largest gain value attribute and $A_0$ and compared with the Hoeffding bound. The value of $\tau$ is used to analyze trade-off conditions. If the attribute obtained is the best choice, then the node is split and the tree grows (lines 13-15) [13], [30], [31].

The lines 21-37 focus on the concept of "transfer learning", where the knowledge gained between two HT models trained on two different signatures (i.e., tasks) are shared or transferred. The model is scalable to transfer knowledge beyond two HT models, if needed. Before we define the transfer learning and how it is used with HT, we must understand the term 'standard learning' (SL). The authors in [29] define SL using the following description process: "Let $D$ be a domain consisting of r-dimensional feature space $X$, a label space $Y$, a probability distribution $P = (x,y)$ where $x \in$ X is the feature vector, and y $\in$ Y. With a finite set of labeled examples $S = \{(x_1, y_1), (x_2, y_2), \dots, (x_n, y_n)\}$ drawn from $P$ and a loss function $J$, standard learning consists of defining a function $f: X \to Y$ with a minimum $J$ value". In the case of transfer learning, a source domain $D_S = (X_S, Y_S, P_S)$ and a target domain $D_T = (X_T, Y_T, P_T)$ are both used to learn a function $f': X_T \to Y_T$ given a set of labeled examples $S$ drawn from $P_T$ and some information pertaining to $D_S$, such that the value of the loss function $J'$ is as small as possible [32]. There are two possible types of transfer learning: inductive and supervised. In the inductive type, the source and target tasks are different, but they share some common features. The supervised type is used when $|S_T| \gg |S_S|$ and aims to improve the task learning of domain $D_T$ given $S_T$ [32]. As shown in Fig.1, the proposed THAT model is trained on four signatures (oscillation and normal events) with different magnitude and durations: 400s, 120s, 60s, and 50s, respectively. Then a supervised transfer learning can be applied to transfer knowledge between different THAT models since $|S_4| \gg |S_3| \gg |S_2| \gg |S_1|$.

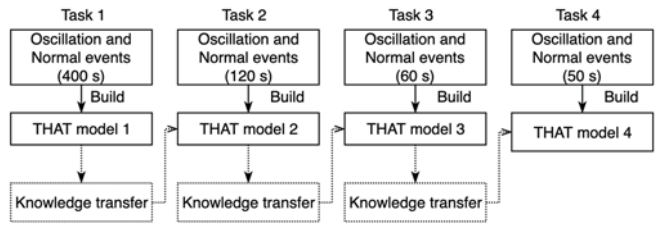

Fig. 1. Supervised transfer learning of THAT model trained on four oscillation events with different durations.

In order to improve the HT, we include the ADWIN as a concept drift detector and make the model adaptable to the eventual concept drifts. ADWIN is an estimator with memory and change detector which is based on the sliding window approach for detecting changes through some statistical tests on different sub-windows. In addition, this window is used to store temporarily the recent data for rebuilding or revising the classifier [28]. The main advantage of ADWIN compared to other sliding window-based approaches is the adaptable window size. Instead of defining a fixed window size, ADWIN uses a dynamic window that adapts its size based on the rate of change of data within the window.

Let $x'_1, x'_2, x'_3, \ldots x'_t$ be a sequence of real PMU values where the value of $x'_t$ is available only at time $t$. Each $x'_t$ was generated based on some distribution $P_t$ and independently for every $t$. Let $u_t$ be the expected value when it is drawn according to $P_t$. ADWIN uses a sliding window $W$ with the recently received data. Let $n$ denote the length of $W$, $\hat{u}_W$ the observed (known) average of the data in $W$, and $u_W$ the expected (unknown) average of the $u_t$ for $t \in W$. If there are two sub-windows $W_0$ and $W_1$ with sufficiently different averages, then it can be concluded that the expected values will be different and the older portion of $W$, i.e. $W_0$, is dropped (Algorithm 2) [33]. The difference between the observed average ($\hat{u}_W$) and the expected average ($u_W$) is compared to $\epsilon$ which is computed based on the Hoeffding bound and is defined as:

$$\epsilon = \sqrt{\frac{1}{2m} \ln \frac{4}{\delta'}} \quad (5)$$

Where $m$ is defined as:

$$m = \frac{1}{1/n_0 + 1/n_1} \quad (6)$$

Where $n_0$, $n_1$ is the lengths of $W_0$ and $W_1$, respectively. The $\delta'$ is defined as:

$$\delta' = \frac{\delta}{n} \quad (7)$$

**Algorithm 2.** ADWIN: Adaptive Windowing Algorithm [33]
1. Initialize Window $W$
2. for each $t > 0$
3.    do $W \leftarrow W \cup \{x'_t\}$ (i.e., add $x'_t$ to the head of $W$)
4.    repeat drop elements from the tail of $W$
5.    until $|\hat{u}_{W_0} - \hat{u}_{W_1}| \geq \epsilon$ holds for every split of $W$ into $W = W_0.W_1$
6. output $\hat{u}_W$

Where $\delta$ is confidence value and $n = n_0 + n_1$.

### B. Dataset

The dataset used in this study includes a collection of oscillatory events (e.g., four signatures) recorded by PMUs across multiple substations at various locations of a power system [34]. Each signature represents a fault in the power grid with associated parameters, e.g., voltage, current, phase angle and frequency information of varying durations, ranging from 3 to 6 minutes. The dataset is modified to introduce concept drift events in the four signatures. Each signature is identified by its oscillation frequency, duration, and the potential cause. For example, a signature containing frequency ranges from 0.1 Hz to 0.15 Hz if holding for up to 400 seconds then it is classified as an oscillation event, otherwise, it is considered as a normal event. The specific range and duration of oscillations have been selected according to the analysis conducted in [18], [35]. Table 1. provides the specifications of the four signatures. To evaluate the adaptability of the uncertain events, the data stream is injected with additional fluctuations or concept drifts using Massive Online Analysis (MOA) generator [30]. A gradual pace of concept drift is modeled to mimic most fault progression in the power system [17]. The final dataset includes four signatures each of which includes 2000 normal events and 2000 oscillation events with gradual concept drifts.

### III. SIMULATION AND RESULTS

With streaming PMU data, it is challenging to store the entire dataset and sectioning it into training, validation, and testing data in order to evaluate the THAT model. Thus, other evaluation techniques are considered including Holdout, interleaved test-then-train, and prequential. In Holdout, a section of the incoming data is used to train the model and small test cases were used to compute the performance. In interleaved test-and-train, each instance of the data stream is used for testing and training the model. Prequential is similar to interleaved test-and-train and uses a sliding window or a decaying factor. The proposed model (THAT) was evaluated using these three techniques and it was observed that the prequential technique exhibited higher detection accuracy with moderate processing time against other approaches.

In order to evaluate the performance of the THAT model, a comparative analysis with OzaBag has been carried out using several performance metrics. An MOA platform [30] was used to run the simulations. The ensemble approach (i.e., OzaBag) is selected, as some literature [21], [36] reported satisfactory results with power system data. The performance metrics used are accuracy, Kappa and evaluation time. The first metric evaluates the algorithm in terms of detecting accurately the instances of oscillation and normal events (Equation (8)). However, it may provide overly prediction results with imbalanced classes. To address this issue, the second metric, (Kappa), can be used to control the instances that may have been correctly classified by chance (Equation (9)). Thus, Kappa is used to help in tuning the THAT model. The evaluation time computes the required time to process each instance.

TABLE I.     PMU DATASET DESCRIPTION

| Signatures | Oscillation frequency | Duration | Potential event cause | Classes | Concept drift |
|---|---|---|---|---|---|
| Signature 1 | 0.1 Hz – 0.15 Hz | $\gg$ 400s | Generators | Oscillation event | Gradual |
| | 0.1 Hz – 0.15 Hz | $\ll$ 400s | - | Normal event | |
| Signature 2 | 0.15 Hz – 1 Hz | $\gg$ 120s | Local plant control | Oscillation event | |
| | 0.15 Hz – 1 Hz | $\ll$ 120s | - | Normal event | |
| Signature 3 | 1.0 Hz – 5.0 Hz | $\gg$ 60s | Inter-area oscillation | Oscillation event | |
| | 1.0 Hz – 5.0 Hz | $\ll$ 60s | - | Normal event | |
| Signature 4 | 5.0 Hz – 14.0 Hz | $\gg$ 50s | Local plant control | Oscillation event | |
| | 5.0 Hz – 14.0 Hz | $\ll$ 50s | - | Normal event | |

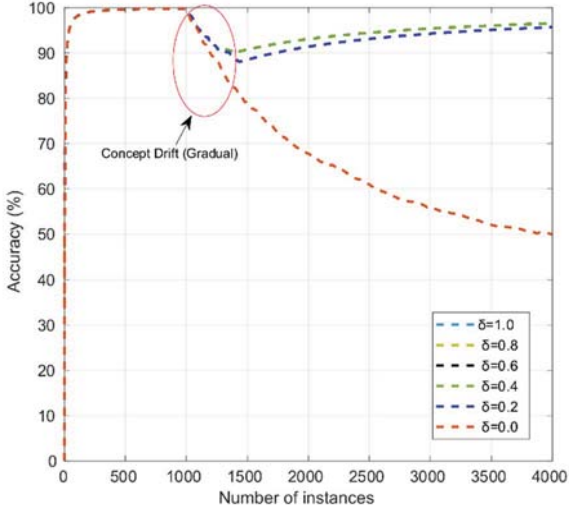

Fig. 2. Accuracy vs Number of Instances for THAT with Gini Index function and different $\delta$ values.

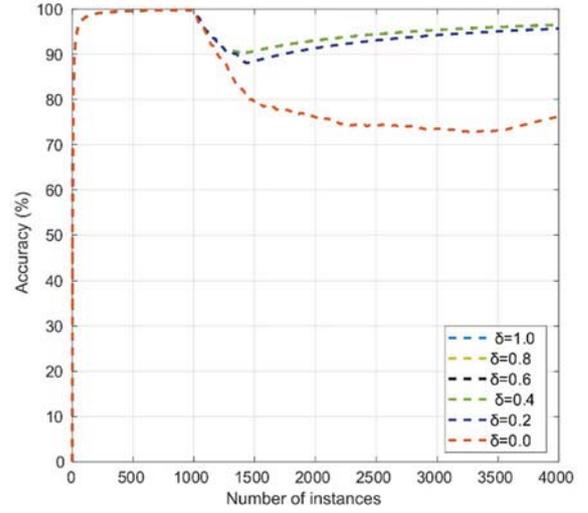

Fig. 3. Accuracy vs Number of Instances for THAT with Information Gain and different $\delta$ values.

$$Accuracy = \frac{TP+TN}{TP+TN+FP+FN} \quad (8)$$

Where $TP$ is the true positive, $TN$ is the true negative, $FP$ is the false positive, and $FN$ is the false negative.

$$Kappa = \frac{(Accuracy - random\ accuracy)}{(1 - random\ accuracy)} \quad (9)$$

Where $Accuracy$ is given by Equation (8), and $random\ accuracy$ is defined as:

$$random\ accuracy = \frac{(TN+FP)*(TN+FN)+(FN+TP)*(FP+TP)}{Total^2} \quad (10)$$

Here $Total$ is $TP + TN + FN + FP$.

### A. Experiment (I): THAT without supervised transfer learning

For this case, THAT model was separately trained on the four signatures without applying supervised transfer learning. A parametric study of THAT and OzaBag models has been conducted in order to define the appropriate value of each hyperparameter of these models. For the THAT, two Information Gain functions have been explored: Gini index and Entropy, and different values of $\delta$ have been selected ranging from 0 to 1. The results of this experiment are given in Figures 2 to 5. These figures illustrate the experimental results of one signature. The results of the other three signatures are not included due to space constraints. Fig. 2 illustrates the accuracy of THAT model as a function of the number of instances with Gini index and different $\delta$ values. As it can be seen, the accuracy increases sharply between 0 and 250 instances to reach the highest accuracy value, which is 99% and stabilizes at that value. The accuracy decreases after the insertion of concept drifts and finally increases again to stabilize at 97%. In addition, one can notice that the model recovers better after gradual concept drift for values of $\delta$ greater than 0.2.

Fig. 3 shows the accuracy of THAT model as a function of the number of instances with the Information Gain and different $\delta$ values. It can be seen that these functions follow the trends similar to those of Fig. 2, the model accuracy increases and reaches the higher accuracy value, which is 99%, and it stabilizes at this value. Then it decreases after the concept drifts and reaches the lowest accuracy value of 72% with $\delta = 0.0$, and then it recovers and stabilizes at 97% after 3500 instances. Additionally, it is worth mentioning that all the highest accuracy values are reported when $\delta$ is set to 0.2 or higher value. As seen in Fig. 2 and 3, THAT + Gini Index and THAT + Information Gain reported the same performance in terms of accuracy especially with a $\delta$ value equal or higher than 0.2. However, the Information Gain is computationally demanding since it uses the logarithmic scale in computing the entropy of each feature. Thus, the Gini Index is selected for the next experiments.

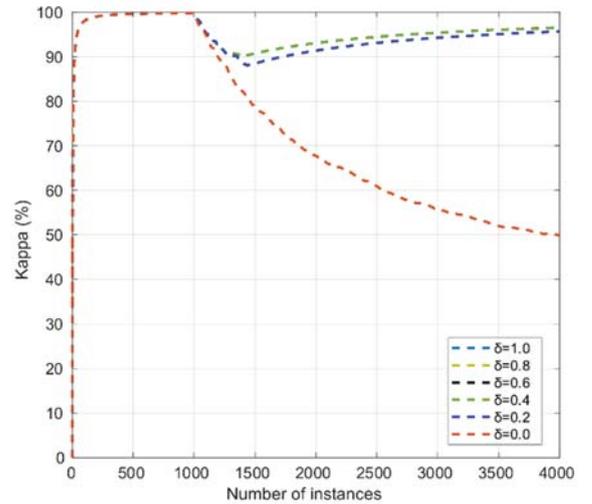

Fig. 4. Kappa vs Number of Instances for THAT with Gini Index function and different $\delta$ values.

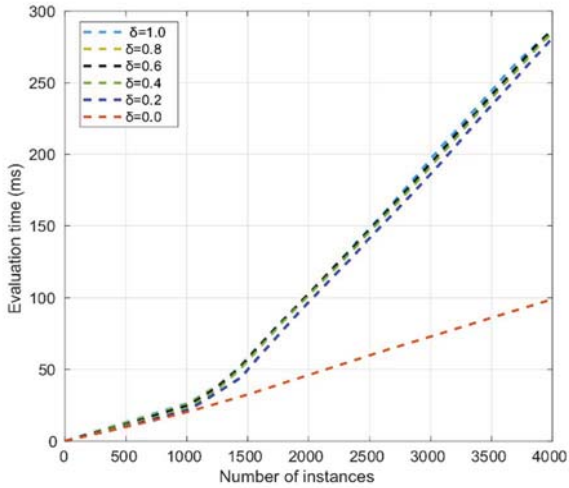

Fig. 5. Evaluation time vs Number of Instances for THAT with Gini Index function and different $\delta$ values.

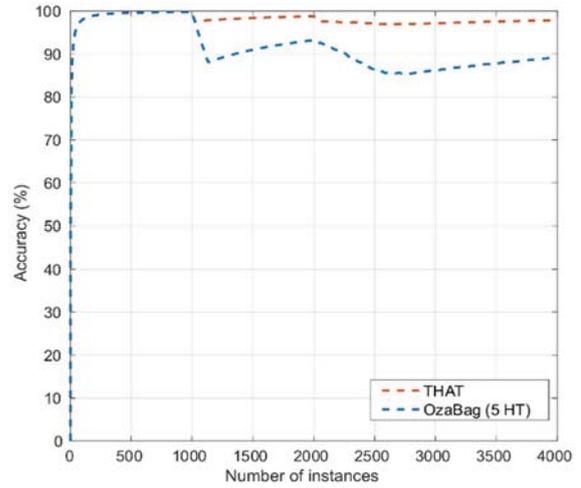

Fig. 6. THAT vs OzaBag in terms of accuracy as a function of the number of instances.

Fig. 4 illustrates the Kappa value of the THAT model as a function of the number of instances with Gini Index and different $\delta$ values. As can be seen, the Kappa value increases exponentially between 0 and 250 instances and stabilizes at 99% and drops at 1000 instances, on the occurrence of concept drifts. After 1000 instances, the model recovers quickly for all $\delta$ values greater than 0.2. Additionally, one can notice that the Kappa function follows a similar trend as that of Fig. 2 since the classes are equally distributed in the dataset (50% oscillation event and 50% normal event).

Fig. 5 illustrates the evaluation time as a function of the number of instances for THAT with Gini Index and different $\delta$ values. It can be seen that the evaluation time increases slightly between 0ms and 25ms with instances less than 1000 instances. After the concept drifts, the evaluation time increases sharply with all $\delta$ values which are equal to or greater than 0.2. The highest evaluation time is reported by $\delta=1$ and the lowest value is reported with $\delta=0$. The reported results suggest that the optimal THAT performance is achieved by Gini Index function and a $\delta$ value which is equal or greater than 0.2. In addition, the THAT model with a $\delta$ greater than 0.4 is slightly demanding in terms of evaluation time. By setting the $\delta$ to 0.2, a tradeoff can be made between high accuracy and optimal evaluation time. Regarding OzaBag, a parametric study has been conducted in terms of the number of HTs: 5, 10, 15, and 20 HTs. The obtained results showed that better performance is achieved by choosing 5 HTs. Increasing the number of HTs does not increase significantly the OzaBag model accuracy but it increases the evaluation time. Thus, OzaBag with 5 HTs is the optimal number in terms of accuracy, Kappa, and evaluation time.

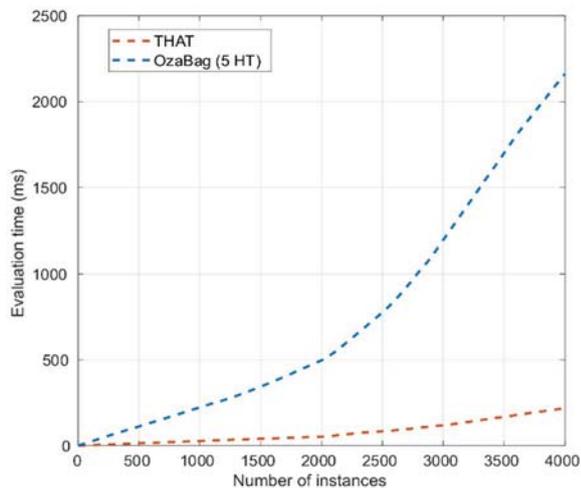

Fig. 7. THAT vs OzaBag in terms of evaluation time as a function of the number of instances.

After defining the appropriate parameters for THAT and OzaBag, the models are trained on the four signatures and the corresponding results are given in Figures 6 through 8. Fig. 6 shows a comparison between THAT and OzaBag in terms of accuracy based on one signature. It is seen that the accuracy of the two models is similar between 0 and 1000 instances. Both instances then reach a peak value of 99% then stabilizes at that value. models' accuracy increases sharply between 0 and 250. After the concept drifts, the accuracy of the models drops to 96% and 88% for THAT and OzaBag, respectively. In addition, one can notice that the THAT model recovers quickly after the concept drift events and increases slightly to reach 98% with 1999 instances. At 2000 instances, the accuracy of THAT model decreases slightly due to another concept drift, but it recovers and regains its previous accuracy value. On the other hand, the accuracy of the OzaBag model is affected by concept drifts. Later, it increases slightly and decreases again after the second concept drift before reaching a final accuracy value of 90% at 4000 instances. Fig. 7 illustrates a comparison between the THAT and OzaBag in terms of the evaluation time. It is observed that the evaluation time of OzaBag increases exponentially as the number of instances increases, while the growth is linear for THAT model. For instance, OzaBag needs 500 ms to process 2000 instances and 2150 ms to process 4000 instances. On the other hand, THAT requires 80 ms for processing 2000 instances and 230 ms for processing 4000 instances. This can be explained by the fact that OzaBag is an ensemble approach which uses several models (e.g., 5 HTs) and training all these models requires longer processing time. However, THAT model only includes one HT model with Gini Index, and hence does not demand larger run-time.

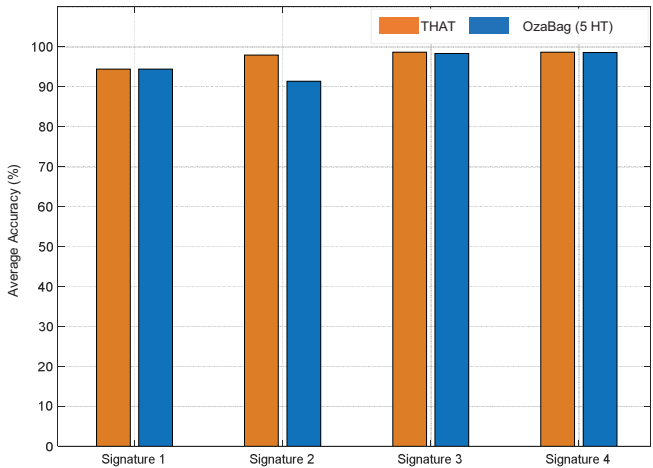
Fig. 8. THAT vs OzaBag in terms of average accuracy

By using the prequential technique, accuracy is computed for each instance. In order to provide an overall accuracy of different models, an average accuracy can be considered which is computed by summing up the accuracy of each instance and dividing the results by the total number of instances. The average accuracy is computed after training the models on the four signatures discussed in Section II and the results are given in Fig. 8. As can be seen, the two models reported similar accuracy, which is 94%, for the first signature. For the second signature, THAT model outperforms OzaBag with a difference of 6%; an accuracy of 97% is reported by THAT while OzaBag's at 91%. For the third signature, THAT model is performing slightly better (99%) than OzaBag (98.5%) For the fourth signature, both models reported an accuracy value of 99%.

### B. Experiment(II): THAT with supervised transfer learning

Here, THAT model has been trained on four signatures using the supervised transfer learning concept. Examples of the obtained results are given in Table II. It is noted that both THAT and the OzaBag models report the same average accuracy of 94%. One can notice that the accuracy of both models has decreased when compared to the first experiment. This is due to the fact that in the first set of experiments, the models have been trained on four signatures separately, while in the second case, these models were trained on four signatures at the same time. Usually, when the models become more generalized, their accuracy decreases. In terms of the evaluation time, THAT model with transfer learning requires a lesser evaluation time than the OzaBag. THAT takes 0.34ms to classify a given instance, while OzaBag takes 1.04ms. A typical PMU with a data rate of 120 samples/s requires 8.33ms to process one sample. Thus, the obtained results suggest that THAT is more suitable for the PMU streaming data in terms of detecting accurately the faults events in near-real-time.

## IV. CONCLUSION

A transfer learning technique using Hoeffding tree and ADWIN is proposed for synchrophasor data. The proposed model, called THAT, can be easily trained for any PMU signatures of varying and shorter durations. It does not require loading the entire data into memory to build the decision tree model, and thus suitable for real-time processing. Additionally, ADWIN is included, so THAT model is easily adaptable to gradual concept drifts. A prequential technique was used to evaluate the performance of the model and the results have been compared to the OzaBag method. After performing a parametric study and tuning the models, two sets of experiments have been conducted. In the first case, THAT model has been trained separately on each signature. In the second case, supervised transfer learning has been applied to THAT model. The obtained results showed that THAT and OzaBag models report higher average accuracy ranging between 91% and 99% (case 1). For case 2, the average accuracy was decreased to 94% in both models, but THAT model required smaller computational run-time than OzaBag. Thus, THAT model is more suitable for the PMU data stream, since it provides a near-real-time response to the dynamic fault event conditions.

TABLE II. COMPARISON BETWEEN THAT MODEL AND OZABAG FOR 4 SIGNATURES (16,000 SAMPLES)

| Datastream models | Average accuracy | Evaluation time/instance |
|---|---|---|
| THAT model with supervised transfer learning | 94% | 0.34ms |
| OzaBag (5 HT) | 94% | 1.04ms |

ACKNOWLEDGMENT

The authors acknowledge the support from the national science foundation (NSF) award # 1537565 for this research work.